\documentclass[]{spie}

\usepackage{amsmath,amsfonts,amssymb}
\usepackage{graphicx}
\usepackage[colorlinks=true, allcolors=blue]{hyperref}
\usepackage{subcaption}
\usepackage{multirow}
\usepackage{xcolor}
\usepackage{longtable}
\usepackage{lscape}
\title{Deep Learning and Large Language Models for Audio and Text Analysis in Predicting Suicidal Acts in Chinese Psychological Support Hotlines}

\author[a]{Yining~Chen}
\author[a]{Jianqiang~Li}
\author[a]{Changwei~Song}
\author[a]{Qing~Zhao}
\author[b,c,d]{Yongsheng~Tong\textsuperscript{*}}
\author[e]{Guanghui~Fu\textsuperscript{*}}

\affil[a]{School of Software Engineering, Beijing University of Technology, Beijing, China}
\affil[b]{Beijing Suicide Research and Prevention Center, Beijing Huilongguan Hospital, Beijing, China}
\affil[c]{WHO Collaborating Center for Research and Training in Suicide Prevention, Beijing, China}
\affil[d]{Peking University Huilongguan Clinical Medical School, Beijing, China}
\affil[e]{Sorbonne Université, Institut du Cerveau - Paris Brain Institute - ICM, CNRS, Inria, Inserm, AP-HP, Hôpital de la Pitié Salpêtrière, Paris, France}

\authorinfo{Further author information: (Send correspondence to Yongsheng Tong: timystong@pku.org.cn and Guanghui Fu: guanghui.fu@icm-institute.org)}

\begin{document} 
\maketitle

%Detailed abstract for technical review purposes (200-300 words).

\begin{abstract}
Suicide is a pressing global issue, demanding urgent and effective preventive interventions. Among the various strategies in place, psychological support hotlines had proved as a potent intervention method. Approximately two million people in China attempt suicide annually, with many individuals making multiple attempts. Prompt identification and intervention for high-risk individuals are crucial to preventing tragedies. With the rapid advancement of artificial intelligence (AI), especially the development of large-scale language models (LLMs), new technological tools have been introduced to the field of mental health. This study included 1284 subjects, and was designed to validate whether deep learning models and LLMs, using audio and transcribed text from support hotlines, can effectively predict suicide risk. 
We proposed a simple LLM-based pipeline that first summarizes transcribed text from approximately one hour of speech to extract key features, and then predict suicidial bahaviours in the future. 
We compared our LLM-based method with the traditional manual scale approach in a clinical setting and with five advanced deep learning models. Surprisingly, the proposed simple LLM pipeline achieved strong performance on a test set of 46 subjects, with an F1 score of 76\% when combined with manual scale rating. This is 7\% higher than the best speech-based deep learning models and represents a 27.82\% point improvement in F1 score compared to using the manual scale apporach alone. Our study explores new applications of LLMs and demonstrates their potential for future use in suicide prevention efforts.  
\end{abstract}

\keywords{Suicide, Psychological support hotlines, Large language model, Deep learning, Speech analysis}

\section{Introduction} \label{sec:intro} 
% 1) Specific position of the question (no general statements on the disease): include a summary of the open scientific questions.
% 2) Purpose of the investigation: what is the specific problem, to what extent the literature tried to solve it, and what is its relevance in the current clinical context?
% 3) How do you plan to solve this problem?

Suicide, a deeply distressing phenomenon, remains prevalent worldwide and has a profound impact on both families and society as a whole~\cite{who2023suicide}. China witnesses an alarming rate of suicides, with an annual average of approximately 287,000 individuals succumbing to it and another two million attempting suicide~\cite{chen2018suicidal}. Recognizing this issue, intervention strategies like psychological support hotlines have been established and proven effective~\cite{middleton2014systematic, mathieu2021systematic}. For instance, Beijing Huilongguan hospital pioneered the introduction of China's first psychological support hotlines, garnering substantial societal benefits~\cite{tong2020prospective}. Nevertheless, a significant challenge persists. Studies show that many people attempt suicide several times before succeeding~\cite{berardelli2020clinical}. Once these individuals are engaged through telephone suicide interventions, the identification of high-risk groups becomes imperative for timely follow-up and intervention. However, the task is daunting due to the overwhelming demands and the large number of individuals in need of support.

In recent years, advancements in AI, particularly in the domain of natural language processing (NLP), have revolutionized many fields~\cite{brown2020language}. AI has been widely applied in suicide detection and prediction~\cite{bernert2020artificial}, with researchers exploring various data types. For instance, Parraga-Alava et al.~\cite{parraga2019unsupervised} used the life corpus containing texts in English and Spanish collected from social networks, blogs, and forums, and proposed a method combining clustering algorithms and semantic similarity measures to categorize potential suicide messages. 
Sawhney et al.~\cite{sawhney2020time} propose STATENet using historical English Twitter context, which is a time-aware transformer-based model for preliminary screening of suicidal risk on social media. Fu et al.~\cite{fu2021distant} proposed a BERT-based distant supervision approach to identify suicide risk on Chinese social media platforms. Qi et al.~\cite{qi2023evaluating} experimented with LLMs for suicide risk prediction on social media data, demonstrating their potential in this domain.
In the context of suicide detection tasks using textual data, most studies have focused on short texts~\cite{parraga2019unsupervised,sawhney2020time,fu2021distant,qi2023evaluating}, which utilized data from SMS text or social media. 
However, few studies have focused on analyzing long texts.
% It would be valuable to investigate whether any research has explored suicide prediction in long-form text. 
Boggs et al.~\cite{boggs2022critical} attempted to predict suicide risk using long-form texts from extensive documents, such as clinical records or patient histories, improving the accuracy of suicide risk prediction models. 
Long text analytics presents unique technical challenges that require further exploration. 
To the best of our knowledge, there is no research on the text analysis using the data from psychological support hotlines.
% While the potential for enhancing predictive performance exists, the complexity and additional resources required may limit the widespread application of such models in clinical settings. Generally, long texts present significant challenges, especially for tasks related to suicide prediction.

Some studies have focused on using speech data for suicidal ideation detection. Scherer et al.~\cite{scherer2013investigating} developed a machine learning model using interview recordings from adolescents to predict future suicidal behavior, Belouali et al.~\cite{belouali2021acoustic} created an AI model using recordings from veterans to predict suicide behavior among U.S. military personnel, and Pillai et al.~\cite{pillai2024investigating} examined the generalizability of speech-based suicidal ideation detection using mobile phone diaries. These studies underscore the significant role of speech data in predicting suicide risk.
However, these studies have limitations. Firstly, the data in these studies are not derived from clinical environments but from interviews~\cite{scherer2013investigating,belouali2021acoustic} or speech recordings related to questionnaires~\cite{pillai2024investigating}. Compared to our study with 1,330 subjects, their sample sizes are relatively small. Sixteen adolescents in Scherer et al.'s study~\cite{scherer2013investigating}, 124 veterans in Belouali et al.'s study~\cite{belouali2021acoustic}, and 786 patients in Pillai et al.'s study~\cite{pillai2024investigating}.
Additionally, our research data is more challenging, with each subject contributing around one hour of speech, while other studies have much shorter durations—up to a maximum of 778 seconds in Scherer et al.~\cite{scherer2013investigating}, and less than three minutes in other studies~\cite{amiriparian2024enhancing, belouali2021acoustic, pillai2024investigating}.
More importantly, the predictive targets in our study (suicidal behavior) were confirmed through follow-up, whereas the targets in other studies were not validated by real-world outcomes but were based on prior suicidal behavior~\cite{scherer2013investigating,belouali2021acoustic} or psychological scales~\cite{pillai2024investigating}. 

In this study, we explore the processing of speech and text data from psychological support hotline to predict suicide risk. We utilized large scale data to build speech analysis models and innovatively employed LLMs for text processing to predict suicidal behavior. By constructing a memory stream, LLMs are able to integrate contextual information from conversations, retain core emotional content related to suicidal tendencies, and significantly reduce data volume without losing critical information. This enables faster and more accurate predictions of suicide risk. Additionally, we investigated a new hybrid approach that combines LLMs with clinical manual scale approach, demonstrating its unique advantages in handling long-term conversations. On a test set containing 46 cases, this proposed LLM pipeline achieved an F1-score of 76.47\%, representing an improvement of 27.82\% points compared to the manual scale rating approach, and around 7.08\% points of F1-score improvement compared to deep learning-based methods. Our study demonstrates an innovative exploration of the potential of LLM in a psychological support hotline suicide prediction task.

\section{Related work} \label{sec:related} 
This study focuses on the application of AI technology in suicide behaviour prediction on psychological support hotlines by analysis the speech data and transcribed text. This section will separately introduce research on suicide related prediction tasks using audio or text data.

\subsection{Suicide related prediction by audio} \label{sec:related:audio}
Some recent studies have integrated computer algorithms for speech analysis in suicide risk prediction by audio data, for example the interview analysis.
Scherer et al.~\cite{scherer2013investigating} developed machine learning models for classifying suicide risk in adolescents, using interviews from 16 teenagers aged 13 to 17. Interview durations averaged 778 seconds for suicidal and 451 seconds for non-suicidal adolescents. The Hidden Markov Model achieved 81.25\% accuracy at the interview level and 69\% at the utterance level.
Amiriparian et al.~\cite{amiriparian2024enhancing} developed an SVM-based machine learning method for detecting suicide ideation using speech data and meta-features. Data from 20 clinical subjects included three speech types: word repetition, picture description, and pronunciation, all under 30 seconds. Meta-information and mental health scales were also collected. A clinical doctor identified suicide ideation on a mental health scale. The model achieved a balanced accuracy of 66.2\% using deep features alone, improving to 94.4\% when combined with meta-information and mental health scales.
Belouali et al.~\cite{belouali2021acoustic} developed machine learning models to predict suicide ideation in U.S. veterans using 588 audio recordings from 124 subjects, averaging 44 seconds each. Suicide ideation was determined through self-report psychiatric scales and questionnaires. Their models, incorporating language and acoustic features, achieved an AUC of 80\%.
Pillai et al.~\cite{pillai2024investigating} developed a model for speech-based suicidal ideation detection using four audio diary datasets collected from mobile phones, involving a total of 786 subjects. The datasets included individuals with major depressive disorder (MDD dataset, 43 subjects), auditory verbal hallucinations (AVH dataset, 356 subjects), persecutory thoughts (PT dataset, 209 subjects), and students with suicidal ideation (StudentSADD dataset, 178 subjects). Audio segments were capped at 180 seconds for the MDD, AVH, and PT datasets, and 30 seconds for the StudentSADD dataset. Suicidal ideation was defined using scales or self-reports. The authors introduced a novel Sinusoidal Similarity Sub-sampling (S3) method to enhance cross-dataset performance, particularly in smaller datasets like MDD, achieving F1-scores of 35\% in MDD, 76\% in AVH, and 83\% in PT during leave-one-dataset-out (LODO) validation.

\subsection{Suicide related prediction by text} \label{sec:related:text}

Another crucial approach of suicide dection is predicting suicide risk through text analysis. 
Tadesse et al.~\cite{tadesse2019detection}  analyze 3549 posts from the SuicideWatch forum to identify suicidal tendencies, and propose an LSTM-CNN combined model, showing that the combined neural network architecture with word embedding techniques achieves the best classification results for suicide ideation. 
Ophir et al.~\cite{ophir2020deep} utilized the Columbia-Suicide Severity Rating Scale (CSSRS)~\cite{posner2011columbia} for evaluating suicide risk, developing a Multi-Task Model (MTM), which integrates multiple risk factors such as texts, personality traits, and psychiatric conditions, demonstrating that combining psychological insights with computational methods greatly enhances suicide risk prediction accuracy. 
Similar, Fu et al.~\cite{fu2021distant} proposed a distant supervision approach based on BERT~\cite{devlin2018bert} to identify suicide risk from Chinese social media.
These studies~\cite{tadesse2019detection,ophir2020deep,fu2021distant} demonstrate the capabilities of AI technology in the field of suicide prediction. 
However, these studies are based on short texts like social media, which cannot be directly applied to the longer transcribed texts from psychological support hotlines used in our study.

% The entire recording from a psychological support hotline typically constitutes long-duration speech, resulting in what is classified as long text once transcribed. Compared to short text, long text is characterized by its richness in background knowledge and logical linguistic structures.
Broadbentet al.~\cite{broadbent2023machine} applies NLP to analyze long text-based crisis counseling sessions. It compares different models and finds that neural network-based models are more effective in detecting suicide risk from long text conversations, highlighting the importance of analyzing extended dialogue to capture nuanced signs of suicidality.
Zang al.~\cite{zang2024accuracy} have utilized long text data, such as patient medical and prescription claims, hospitalization records, and various types of visit documentation, including outpatient, inpatient, and emergency visits. They structured this extensive textual information into organized electronic health records (EHRs). Several published algorithms, developed using data mining and machine learning methods, have been employed to predict suicide attempts and suicide mortality within large healthcare systems. However, current research on long text and long audio for suicide detection is relatively limited, for instance Zang et al.~\cite{zang2024accuracy} focus on structured data, such as electronic health records extracted from clinical records or patient histories. Some research, like that of Broadbent et al.~\cite{broadbent2023machine}, has encountered limitations, with models showing high false positive rates, reaching up to 75\%. Generally, long texts of unstructured data present significant challenges, particularly for tasks related to suicide prediction.

\subsection{LLM in medical domain} \label{sec:related:llm}
After the great success of Transformer-based language models such as BERT~\cite{lee2018pre} and GPT~\cite{radford2018improving}, researchers and practitioners have advanced towards larger and more powerful language models (e.g., OpenAI’s ChatGPT~\cite{zhao2023survey}). These instruction-finetuned Large Language Models (LLMs), including GPT-4~\cite{gpt4}, PaLM~\cite{palm}, and LLaMA~\cite{llama} , boast tens to hundreds of billions of parameters and demonstrate emergent abilities that significantly outperform those of their smaller, pre-trained counterparts. Initially conceived for understanding and generating human-like text, LLMs have found diverse applications for example medical report assistant~\cite{jeblick2022chatgpt}, and answering medical related questions~\cite{yeo2023assessing}. 
Additionally, this technique has garnered significant attention in the mental health domain~\cite{he2023towards,farhat2023chatgpt}, with numerous applications being developed in this field.

Yang et al.~\cite{yang2023towards} have conducted a comprehensive evaluation of ChatGPT's capabilities in psychological health analysis and emotional reasoning across eleven datasets covering five tasks. These tasks include binary and multiclass mental health state detection, identifying causes or factors behind mental health states, recognizing emotions in conversations, and analyzing causal emotional implications, confirming ChatGPT's superiority over traditional neural network methods. Additionally, Bhaumik et al.~\cite{bhaumik2023mindwatch} have developed MindWatch—a classifier aimed at detecting suicidal intentions. It leverages the ALBERT and Bio-Clinical BERT language models, fine-tuned with data from Reddit, to provide customized psychological educational plans for individuals facing mental health challenges. By delivering tailored content specific to unique mental health conditions, treatment options, and self-help resources, this method encourages individuals to actively engage in their recovery process. Further, Xu et al.~\cite{xu2024mental} have extensively assessed the performance of various LLMs in predicting mental health states from online social media text data, conducting experiments with zero-shot, few-shot, and guided fine-tuning approaches. The results indicate that guided fine-tuning significantly improves the LLMs' ability to address specific mental health-related tasks across multiple datasets. These studies demonstrates LLMs's strong contextual learning abilities in mental health , which is able to generate explanations close to human performance.

We used LLMs to process the long transcribed texts from psychological support hotlines, which are approximately one hour in length in our study. 
The first step is to summarize the main information related to suicide. Many studies have validated the performance of LLMs in text summarization within the medical domain. For example, Van Veen et al.~\cite{vanveen2023clinical} explored the application of LLMs in clinical text summarization, covering various tasks such as summarizing radiology reports and doctor-patient dialogues. They experimented with eight LLMs across six datasets and four summarization tasks. Their experiments demonstrated that LLM-generated summaries might outperform humans in certain clinical tasks, allowing clinicians to focus more on patient care by relying on LLMs for documentation. 
However, due to the nature of LLMs, there is a risk of hallucination, where the model may generate content that is not based on real or accurate information~\cite{tonmoy2024comprehensive}.
Tang et al.~\cite{tang2023evaluating} found that GPT-3.5 and ChatGPT can produce factually inconsistent medical summaries, highlighting a gap between automatic metrics and true summary quality, raising concerns about misinformation.
This suggests that we need a solution to enable the LLM to summarize content in a way that avoids disjointed or inaccurate information, which is especially important in tasks involving long texts.

\subsection{Comparison with related studies}\label{sec:related:summary}
In our study, we utilized the largest dataset from a clinically used psychological support hotline, with the gold standard confirmed through 12-month follow-ups. 
We analyzed the data from both speech audio and transcribed text using deep learning models and LLMs separately.
Our study included 1,284 subjects, a significantly larger sample size compared to related research, such as Scherer et al.'s 16 adolescents~\cite{scherer2013investigating}, Belouali et al.'s 124 veterans~\cite{belouali2021acoustic}, and Pillai et al.'s 786 patients~\cite{pillai2024investigating}.
More importantly, our approach offers greater reliability by validating actual suicidal behavior through follow-up confirmation rather than relying on past behavior~\cite{scherer2013investigating,belouali2021acoustic}, psychological scales~\cite{amiriparian2024enhancing, pillai2024investigating}, self-report suicidal thoughts~\cite{belouali2021acoustic}, or data from social media that are not easily identifiable as indicators of suicide risk~\cite{tadesse2019detection, fu2021distant}.
From the audio analysis part, our dataset includes long-duration audio recordings averaging 59.6 minutes, which posed unique challenges for model design. Unlike the shorter audio lengths in the other studies (up to 778 seconds in Scherer et al.~\cite{scherer2013investigating}, around 30 seconds in Amiriparian et al.~\cite{amiriparian2024enhancing}, 44 seconds in Belouali et al.~\cite{belouali2021acoustic}, and 180 seconds in Pillai et al.~\cite{pillai2024investigating}), we employed advanced methods tailored to our data, as traditional models like Hidden Markov Models and SVMs may unsuitable for our task. 
From the perspective of text processing for suicide prediction, we utilized an LLM driven by a few-shot samples, which did not require additional training. 
To the best of our knowledge, there has been no prior research on our topic that utilized data from a clinical psychological support hotline to predict future suicide behaviour.

\section{Processes of the Beijing psychological support hotline} \label{sec:background}

The psychological support hotline was established at the end of 2002, located at Beijing Huilongguan Hospital and the Beijing Suicide Research and Intervention Center. It operates 24 hours a day, 365 days a year, and is staffed by certified psychological operator from the Beijing Suicide Research and Intervention Center. The hotline provides mental health education to the public, psychological support to callers in crisis, reduces suicide risk for high-risk callers, offers information on mental health, encourages those with psychological issues to seek professional treatment, and provides referrals to mental health institutions. 
The operation of the psychological support hotline can be divided into three stages: suicide risk assessment, crisis intervention, and follow-up support. We provide an introduction to these three stages.
For simplicity, the staff operating the psychological intervention hotline will be referred to as ``Operator'', and individuals calling the hotline will be referred to as ``Caller''. All \textgreater 30 hotline operators were trained in assessing suicidal risk before answering calls independently. Details of the process can be seen in references~\cite{tong2020prospective, tong2023predictive}. The workflow of the psychological support hotline can be seen in Figure~\ref{fig:flowchart_full}~(A).

\subsection{Suicide risk assessment} \label{sec:background:assessment}
Based on the emotional state and factual situation exhibited by callers, hotline operators will use the suicide risk assessment scale developed by Tong et al.\cite{tong2020prospective} to objectively evaluate the caller's suicide risk. This assessment scale consists of 12 factors, including suicidal ideation and plans, severe depression, acute life stress, chronic life events, alcohol and drug abuse, among others. The individual scores from the suicide risk assessment scale are summed, resulting in a total score ranging from 0 to 16. A score of 8-16 indicates high risk, while a score of 0-7 indicates low-moderate risk. Incomplete data from respondents failing to answer more than five items results in missing scores. 
Typically, it takes 10-15 minutes to complete the scale. The items evaluated in the suicide risk assessment scale can be seen in Table~\ref{tab:scale}. 
Subsequently, the crisis intervention phase focuses on immediate risk mitigation and support. Ideally, the scale should be administered first, followed by crisis intervention. However, if the caller is exhibiting high suicidal ideation, the operator must provide immediate crisis intervention or be flexible in the sequence, incorporating scale questions as needed.

\begin{table}[!ht]
\begin{center}
\caption{Suicide risk assessment scale used in Beijing psychological support hotline, developed by Tong et al.~\cite{tong2020prospective, tong2023predictive}.}
\label{tab:scale}
\begin{tabular}{l c}
\hline
\textbf{Elements (number of items) } & \textbf{Score} \\
\hline
Suicidal ideation and plan (3) & 0/1/4 \\
Severe depression (11) & 0/1 \\
Hopelessness (1) & 0/1 \\
Psychological distress (1) & 0/1 \\
Acute life events (2) & 0/2 \\
Chronic life events (2) & 0/1 \\
Alcohol or substance misuse (3) & 0/1 \\
Severe physical illness (1) & 0/1 \\
Fear of being attacked (2) & 0/1 \\
History of being abused (2) & 0/1 \\
Suicide attempt history (1) & 0/1 \\
Relatives or acquaintances suicidal acts history (2) & 0/1 \\
\hline
\end{tabular}
\end{center}
\end{table}

\subsection{Crisis intervention} \label{sec:background:intervention}

The Beijing Psychological Crisis Research and Intervention Center developed a semi-structured crisis hotline intervention better suited to Chinese individuals at high risk. Operators are trained to show empathy and respect, and to provide emotional support to callers. After assessing the caller's suicide risk, operators focus on non-directive and collaborative problem-solving and explore feasible strategies for coping with suicide risk. For callers at imminent risk of suicide, operators employ various strategies, including developing a safety plan with the caller; contacting the caller's family, friends, or the police; or establishing a no-suicide agreement.

\subsection{Follow-up support} \label{sec:background:followup}
At the end of the call, the operator would discuss the follow-up call schedule: one day, one week, one month, and three month after the intake call. In each scheduled follow-up, operators continue to provide emotional support, inquire about suicidal thoughts, and offer psychological interventions if necessary. To evaluate the incidence of suicidal behavior during the follow-up period, operators ask, ``Since [the date of the intake call], have you attempted suicide?'' If the caller cannot be reached, the operator will ask the family members who answer the follow-up calls if the caller has engaged in suicidal behavior and will set up a new follow-up schedule for those still alive. If the caller cannot be reached after three attempts at each scheduled follow-up point, they will be marked as ``lost'', and the operator will try to reach them again at the next scheduled follow-up, continuing until either the caller's death or 90 days after the intake call.

\section{Methods} \label{sec:methods}
\begin{figure}[!hbtp]
\centering
\includegraphics[width=1\linewidth]{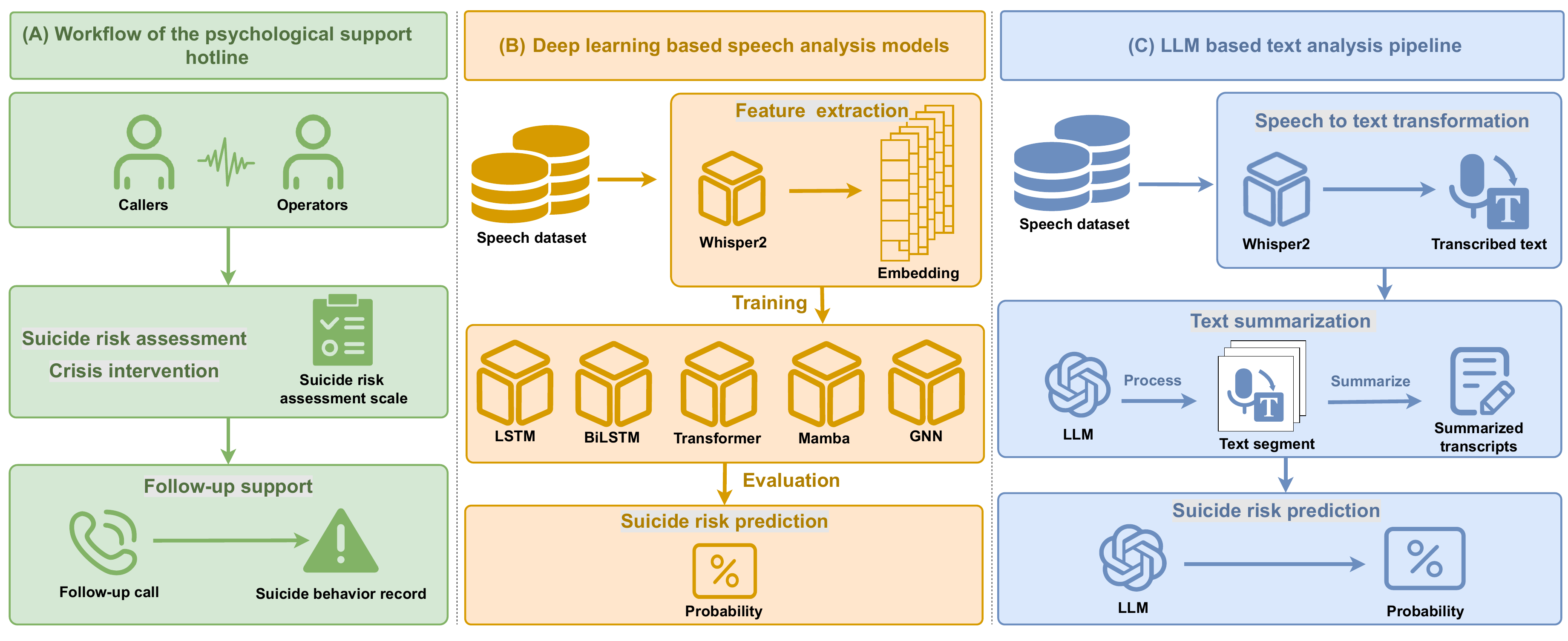}
\caption{The flowchart of this research, including (A) the workflow of the psychological support hotline, and two parts of experiments for the suicidal prediction task: (B) deep learning based speech analysis models, and (C) the proposed LLM text analysis pipeline.}
\label{fig:flowchart_full}
\end{figure}

The workflow of our study on method development can be seen in Figure~\ref{fig:flowchart_full}, with (B) detailing the audio analysis and (C) detailing the text analysis separately.

% 待修改

\subsection{LLMs based text analysis} \label{sec:methods:llm}
\subsubsection{Speech to text transformation}

In the process of converting speech to text, we employ the Whisper2~\cite{radford2023robust}  model for the transcription task.
Whisper2, developed by OpenAI, is an advanced speech recognition technology. Firstly, the model has been trained on an extensive and diverse dataset, enabling it to accurately recognize speech across various accents and linguistic contexts. This comprehensive training allows Whisper2 to adeptly handle subtle sound variations and dialects, ensuring its effectiveness in different language environments. Furthermore, Whisper2 excels in multilingual support. It can recognize not only standard language forms but also a wide range of colloquial expressions and dialects. Additionally, Whisper2 demonstrates significant robustness when dealing with varying speech quality and environmental noise. This versatility makes Whisper2 an ideal solution for applications requiring reliable speech recognition across multiple linguistic contexts, addressing the demands of complex, multi-context language processing.
We utilized the offline text-to-text model Whisper2 to convert the raw speech recordings into a transcribed text $D_t$.

\subsubsection{Context-based text summarization for memory stream conduction}

We employ a memory stream conduction to summarize the caller's emotional state and potential suicidal tendencies, aiding the LLM in forming a chain of thought. In the process of constructing memory streams for long psychological support hotline conversations, we segment complex dialogue texts into more manageable small segments. For each segment, we rely on three main components to retrieve memory backgrounds and form new comprehensive summaries: Recency, Importance, and Relevance. 
\begin{itemize}
    \item Recency refers to prioritizing recently accessed memory content, thereby making the most recent events more easily retrievable. For instance, this includes considering the patient's current psychological problems and their emotional state. 
    \item Importance involves distinguishing between routine events and key memories, assigning higher priority to more significant events. Examples include the patient’s depressive tendencies, clearly expressed suicidal intentions, and the current level of suicide risk. 
    \item Relevance means assigning a higher weight to memories related to the current situation. This includes factors such as changes in the patient's emotional state, the degree of self-focus, and the characteristics of stress.
\end{itemize}
Ultimately, the memory stream generates a comprehensive summary that is not merely a simple compilation of previous memories but a complete record of the caller's recent condition and tendencies toward suicide.

We employed ChatGLM2-6B~\cite{glm2024chatglm} for the offline small-scale LLM, aimed at ensuring the privacy and security of the data. ChatGLM2-6B is an open-source dialogue language model developed by Tsinghua University that supports both Chinese and English. After pre training with 1.4T Chinese English identifiers and human preference alignment training, ChatGLM2-6B with 6.2 billion parameters has been able to generate answers that are quite in line with human preferences. By combining model quantization technology, we can deploy locally on consumer grade graphics cards (with a minimum requirement of 6GB of graphics memory at INT4 quantization level). By utilizing this offline model, we could efficiently process sensitive speech-to-text data while adhering to strict confidentiality requirements. The ChatGLM2-6B model's robust capabilities allowed us to handle the complex tasks of language processing without compromising the integrity and secrecy of the information processed.

\begin{figure}[!hbtp]
\centering
\includegraphics[width=1\linewidth]{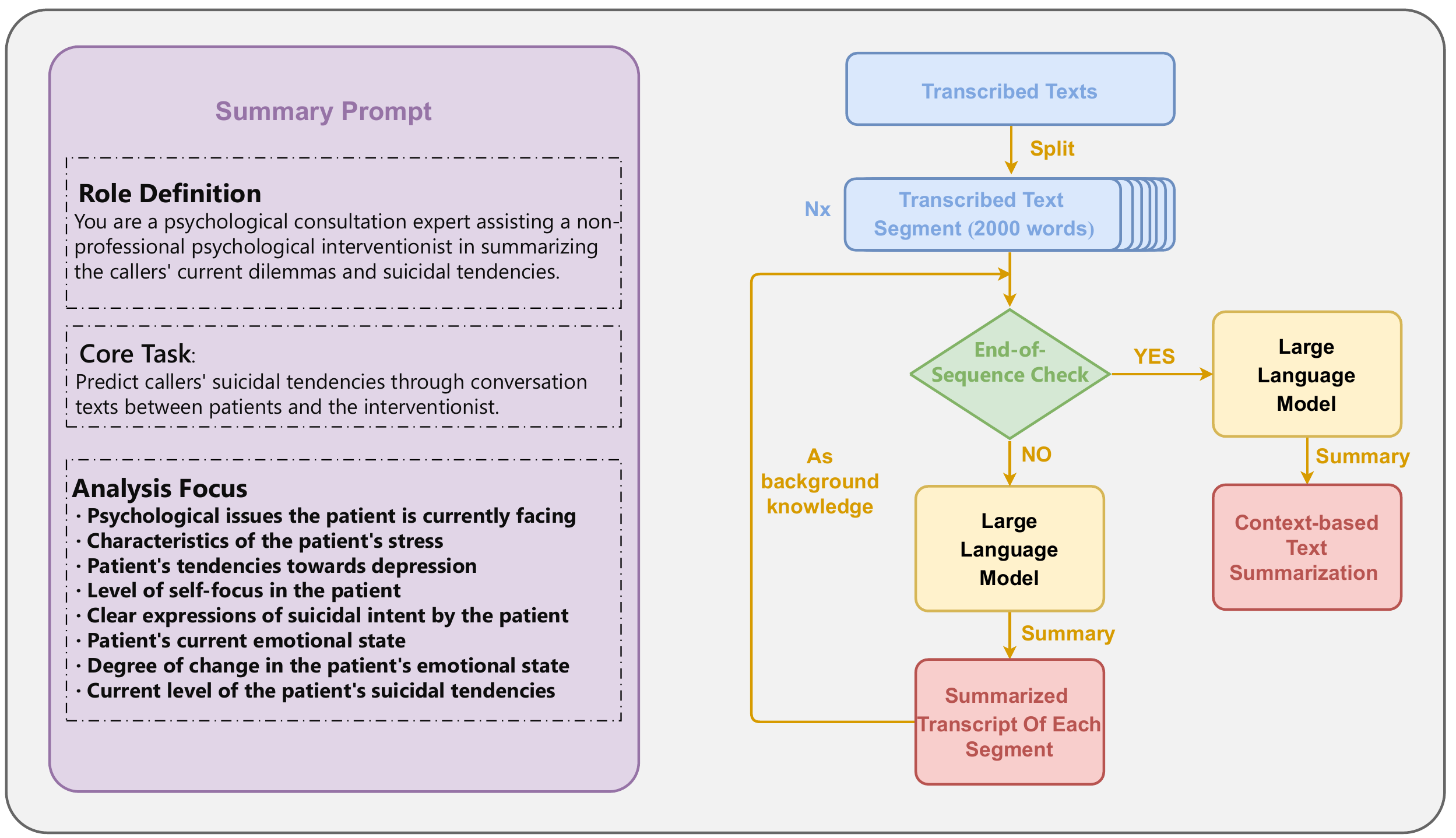}
\caption{Process and Details of Generating Methods for Segmentation and Summarization. }
\label{fig:SummarizedTranscript}
\end{figure}

In the first step, the transcribed text $D_t$ is fragmented into multiple segments, from $d_1$ to $d_n$, to fit within the text limitation of the utilized LLM.
For every segment $d_i$, we employed the LLM to generate a summarized text. The current processing segment also relied on the previous segment $d_{i-1}$ to maintain the memory stream, and the output summarization is $s_i$.
Finally, we collected all these summaries $S$ as the input for the next step. The detailed prompt for text summarization is shown in Figure~\ref{fig:SummarizedTranscript}.

\subsubsection{LLM for suicide prediction}

\subsection{Deep learning based speech analysis}\label{sec:methods:deep_learning}

In the speech analysis part, we first segmented the speech data from the psychological support hotline into 30 second intervals. Then, we used Whisper2~\cite{radford2023robust} as a feature extractor to convert each segment into a set of feature embeddings, which served as the input for the decoder in the suicide prediction task.
We selected five popular deep learning decoders for suicide prediction task, as decribed below:

\paragraph{LSTM:~\cite{hochreiter1997long}}Long Short-Term Memory (LSTM) networks are a type of recurrent neural network (RNN) designed to model sequential data by capturing long-term dependencies and patterns. When applying LSTM to suicide prediction using speech features, the speech data is treated as a sequence of feature vectors. LSTM processes this sequence step-by-step, maintaining a memory cell that preserves relevant information over long sequences. The network selectively updates, forgets, or retains information in the memory cell through gated mechanisms, allowing it to capture both short-term and long-term dependencies in the speech signal. By learning these temporal patterns, LSTM can effectively identify subtle cues and trends in the speech features that may be indicative of suicidal tendencies, thereby improving the prediction accuracy and reliability.
\paragraph{BiLSTM:~\cite{graves2005bilstm}} Bidirectional Long Short-Term Memory (BiLSTM) networks enhance the capabilities of standard LSTM by processing sequential data in both forward and backward directions. When applying BiLSTM to suicide prediction using speech features, the speech data is treated as a sequence of feature vectors. The BiLSTM network has two parallel LSTM layers: one processes the sequence from start to end, while the other processes it from end to start. This bidirectional approach allows the model to capture information from both past and future contexts for each point in the sequence. By integrating information from both directions, BiLSTM can more effectively capture complex temporal patterns and dependencies within the speech features. This results in a more comprehensive understanding of the speech data, enhancing the model's ability to detect subtle cues and trends associated with suicidal tendencies, thereby improving prediction accuracy and reliability.

\paragraph{GNN:~\cite{GNN}} The principle of applying Graph Neural Networks (GNNs) to suicide prediction using speech features leverages GNNs' powerful ability to model graph-structured data. In this task, speech frame level features are treated as nodes, where each node represents an speech segment or feature point. GNN constructs a graph structure by connecting similar speech features, forming edges between nodes. Then, GNN iteratively updates node representations by aggregating information from neighboring nodes. This information propagation mechanism ensures that each node representation not only contains its own feature information but also integrates the contextual information of its neighbors, capturing complex dependencies and global patterns in the speech data. This approach effectively utilizes the relationships among speech features.

\paragraph{Transformer:~\cite{vaswani2017attention}} The Transformer model is a neural network architecture designed to handle sequential data using self-attention mechanisms, which enable efficient parallel processing. When applying the Transformer to suicide prediction using speech features, the speech data is first transformed into a sequence of feature vectors. The Transformer consists of an encoder and a decoder. The encoder processes the input sequence through multiple layers of self-attention and feed-forward neural networks, capturing global dependencies and relationships within the speech signal. The decoder takes the encoded representation and generates the final output, allowing the model to focus on relevant parts of the sequence during prediction. This comprehensive approach ensures that the Transformer can effectively identify complex patterns and subtle cues in the speech features, improving the accuracy and reliability of suicide prediction.

\paragraph{Mamba:~\cite{gu2023mamba}} Mamba is a deep learning model based on the State Space Model (SSM), aimed at improving performance and computational efficiency for long sequence tasks. It processes input data sequentially, using a stack of multiple MambaBlock modules to capture complex patterns and relationships in the data. The design of the Mamba model draws inspiration from traditional state space models, enhancing computational efficiency through discretization and convolutional methods. Additionally, the Mamba model introduces selective mechanisms and hardware-aware algorithms to further optimize computational efficiency and performance.

\section{Experiments} \label{sec:experiments}

\subsection{Datasets} \label{sec:experiments:dataset}

This study was supported by the National Natural Science Foundation of China [82071546], Beijing Municipal High Rank Public Health Researcher Training Program [2022-2-027], the Beijing Hospitals Authority Clinical Medicine Development of Special Funding Support [ZYLX202130], and the Beijing Hospitals Authority’s Ascent Plan [DFL20221701]. 
All the data were taken from the psychological support hotlines of Beijing Suicide Research and Prevention Center, Beijing Huilongguan Hospital from 2015-2017. 
Prior to connecting to a crisis line counselor, callers were informed by voice message that all calls are taperecorded and that data about the call would be collected and analyzed anonymously. The identity of the callers was deidentified before data analyses.
We included a total of 1,284 subjects (42\% males and 58\% females), aged 11-67 years (mean 26, $\pm$ 7.80). 
We selected the speech for complete follow-up interviews. Cases of suicide were confirmed through follow-up interviews.
Our research includes two part of models: one based on few-shot prompts for LLMs which only require few training data as prompt, and the other deep learning models that require large-scale training data. 
The data used for LLM prompt construction includes 6 subjects through follow-up interviews (3 males and 3 females), aged 23-30 (mean 26, $\pm$ 4.05) comprising three subjects with suicidal behavior and three subjects without suicidal behavior.

For deep learning model conduction, we used a total of 1232 subjects. 
The durations of these speech range from 9 minutes to 120 minutes, with a total duration of approximately 1538.6 hours, averaging around 59.6 minutes. 
We split into training and validation sets at a ratio of 4:1. The training set containscontains 985 (42\% male, 58\% female) aged 11-67 (mean 26.4, $\pm$ 8.03). The validation set contains 247 subject (45\% male, 55\% female), aged 12-55 (mean 26.1, $\pm$ 8.13). Among the training set, 460 subjects had committed suicide, while the remaining 525 did not exhibit suicidal behavior. For the validation set, 125 subjects had committed suicide, while the remaining 122 did not exhibit suicidal behavior. 

\begin{figure}[!hbtp]
\centering
\includegraphics[width=1.0\linewidth]{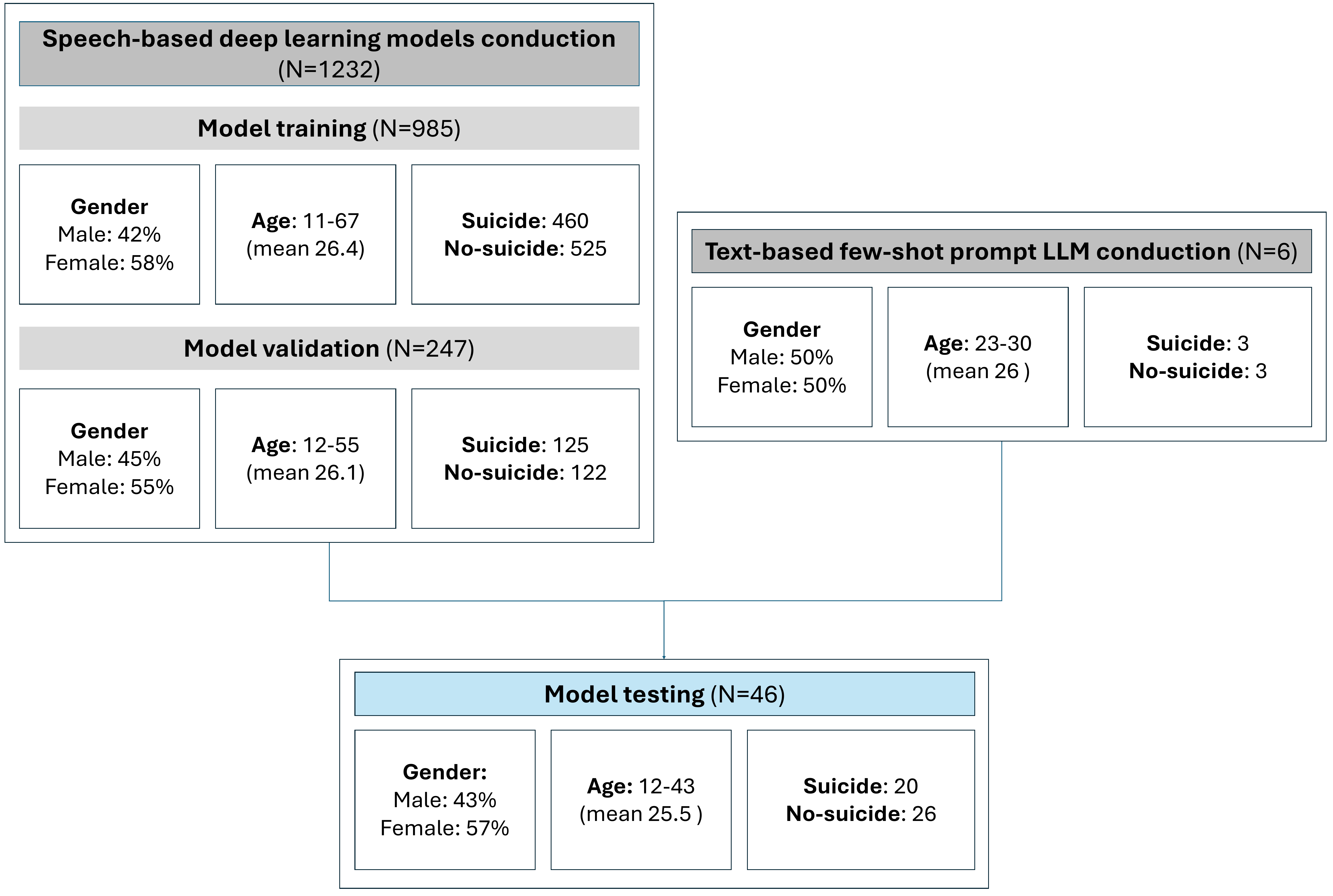}
\caption{Data distribution for deep learning model training and validation, and few-shot prompt LLM conduction. $N$ denotes the number of subjects.}
\label{fig:flow_chart}
\end{figure}

Based on the availability of data, we selected 46 callers (43\% male, 57\% female) aged 12-43 (mean 25.5,$\pm$ 7.01) as the test set, of which 20 exhibited suicidal behavior, while the remaining 26 did not. These individuals have complete scale score records, detailed follow-up interviews, and fully available psychological support hotline recordings. The test sets were only used to evaluate the performance of the models. 
The detailed flowchart and data distribution for this research can be seen in Figure~\ref{fig:flow_chart}.

\subsection{Experiment design and implementation details}\label{sec:experiments:experiment_design}

\subsubsection{LLM pipeline} \label{sec:experiments:experiment_design:llm}
Due to data and privacy protection issues, we cannot directly use the online version of LLM. Due to limitations in the hospital experimental environment and the lack of support for deploying larger-scale LLMs, we select ChatGLM-6B for text summarization. 
For the entire speech text, we split the input into segments of 2000 Chinese characters for summarization, with the output not exceeding 512 Chinese characters.
The summaries generated by ChatGLM2-6B underwent data anonymization to ensure privacy. 
The output after text summarization no longer contains any identifiable information, ensuring privacy and security. Therefore, we can use a larger LLM for the next step of suicide prediction. In our experiment, we used GPT-4 for this step. 
For prompt conduction, we employed two approaches: zero-shot and few-shot, as described in Section~\ref{sec:methods:llm}.
Additionally, we experimented with combining the LLM approach with the traditional manual scale rating method. 
In this approach, the weights for the manual scale and the LLM output were set as $\alpha = 0.5$ and $\beta = 0.5$, respectively.

\subsubsection{Deep learning methods}
\label{sec:experiments:experiment_design:deep_learning}
All experiments were conducted on an NVIDIA GeForce RTX 4090 with 12GB of GPU memory. The models were developed using the PyTorch\cite{paszke2019pytorch} framework. We employed a total of five advanced deep learning models for the suicide prediction task as introduced in Section~\ref{sec:methods:deep_learning}.
All these models were trained 500 epochs, with a batch size of 32, using the AdamW~\cite{kingma2014adam} optimizer. 
As mentioned before, we used Whisper-large-v2 model as the feature extractor. Each speech segment was converted into 1280-dimensional embeddings. 
We did not perform any additional preprocessing.

For evaluation, we used sensitivity, specificity, precision, and F1-score as metrics for our task. For each metric, we reported its mean value as well as the corresponding 95\% confidence interval (CI) computed using bootstrap over the independent test set. 
The source code of all these methods are public avaiable at: \url{https://github.com/xiaoning317/Suicide_Risk_Prediction}.

\subsection{Explainability of LLM} \label{sec:experiments:explainability}

We utilized LLM to perform summarization and analysis on long text records from the psychological support hotline, aiming to predict the suicide risk of callers. In this task, explainability is achieved by not only having the model generate prediction scores but also requiring it to provide the rationale behind these scores~\cite{yang2023towards}. We involved experienced operators and professional psychologists from Beijing Huilongguan Hospital to analyze and interpret the model's output. After evaluation by these experts, the validity of using summaries to predict suicide risk was confirmed, demonstrating the model's capability to perform this task reasonably.

\section{Results}\label{sec:results}
The proposed LLM pipeline achieved a Dice score of 75 [59.08, 86.79], while the best-performing deep learning model Mamba achieved a Dice score of 69.39 [53.33, 81.63] on the test set (in \%, mean [95\% bootstrap CI]).
Table~\ref{tab:results} presents the performance of various models using different modalities of data for the suicide prediction task using the four evaluation metrics (sensitivity, specificity, precision, and F1-score). Through comparisons with traditional deep learning models, we find that few-shot LLM pipeline demonstrate superior performance in predicting callers' suicide probabilities from text. It is worth noting that when combined with the manual scale rating, the performance improved to an F1 score of 76.47\%, which is 1.47\% points higher than the LLM alone and 27.82\% points higher than the manual scale approach. 
The performance of deep learning models based on speech analysis is also strong compared to the manual scale rating.
GNN and Mamba perform the best in balancing sensitivity and specificity, while also having similar high F1-scores. 

\begin{figure}[!hbtp]
\centering
\includegraphics[width=1\linewidth]{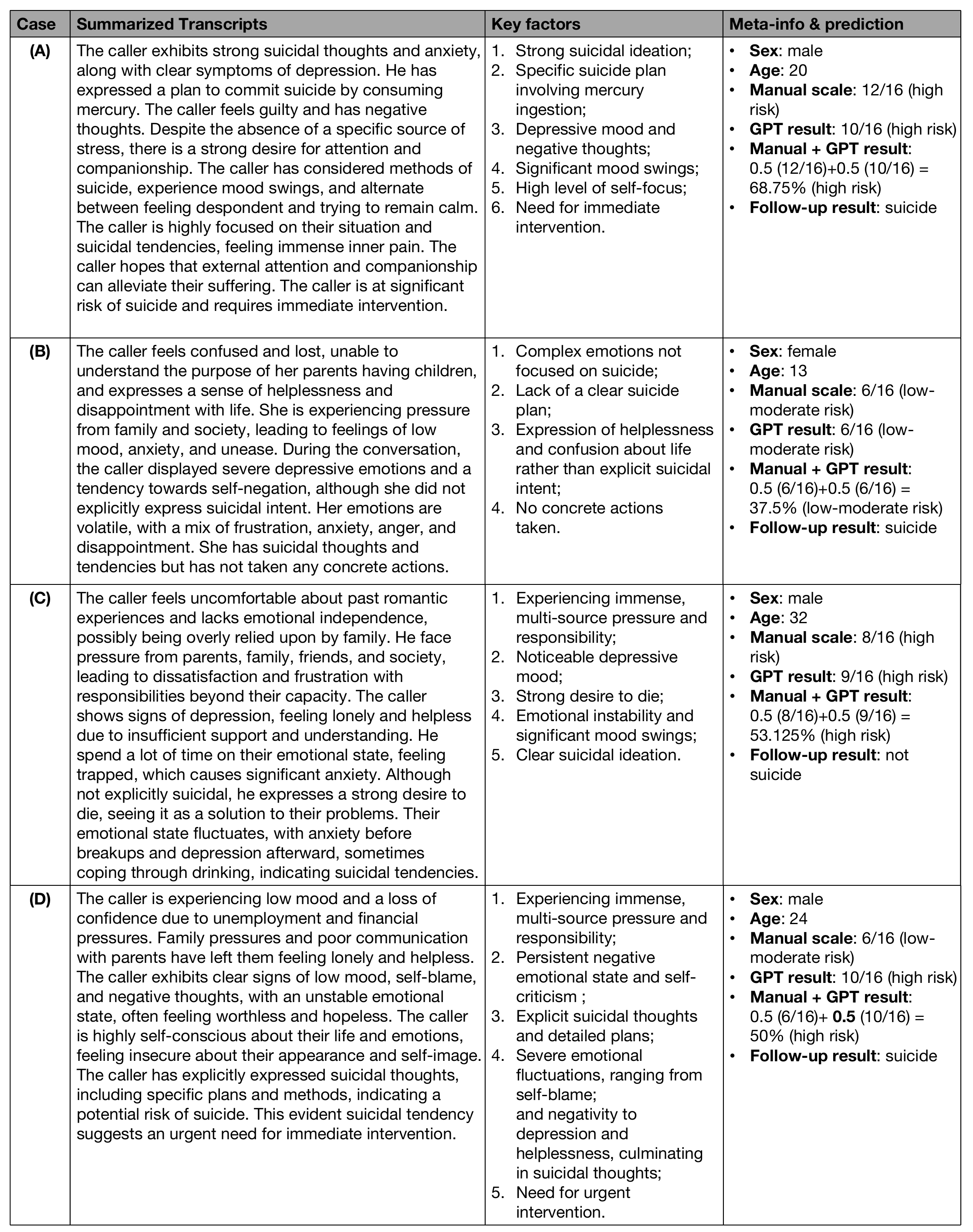}
\caption{The summarized transcripts and key factors related to the final decision generated by the LLM, along with the meta-information (including the ground truth) of case (A), (B), (C) and (D) and model predictions. Note that, The GPT output is not a direct mental scale score but is instead designed to align with the manual scale scoring system for combined manual+GPT calculations. ``High risk'' refers to a high risk of suicide. }
\label{fig:example}
\end{figure}

We present four detailed examples in Figure~\ref{fig:example}, including summarized text, key factors output by the LLM, and the meta-information along with the model prediction.

\begin{table}[!ht]
\centering
\caption{Performance comparison of traditional manual scale rating and five deep learning methods and three LLM pipelines on suicide prediction task. The results are presented as mean with 95\% bootstrap confidence interval computed on the test set contains 46 subjects.}
\label{tab:results}
\resizebox{1\linewidth}{!}{
\begin{tabular}{|l|l|l|l|l|l|} 
\hline
Data                    & Methods                   & Sensitivity         & Specificity          & Precision           & F1-score              \\ 
\hline
\multirow{7}{*}{Speech} & Manual scale rating 
                        & 60.00~[35.71,85.00]      
                        & 40.91~[20.00,62.50]      
                        & 40.91~[22.22,61.90]      
                        & 48.65~[28.57,66.67]
                        \\ 

\cline{2-6}
                        & LSTM            
                        & 95.00~[83.00,100.00]      
                        & 30.77~[14.29,48.15]      
                        & 51.35~[36.11,67.65]      
                        & 66.67~[51.85,80.00] 
                        \\ 
                        
\cline{2-6}
                        & BiLSTM                
                        & 85.00~[68.42,100.00]      
                        & 46.15~[27.26,65.42]      
                        & 54.84~[37.50,72.41]      
                        & 66.67~[50.00,80.00] 
                        \\ 
                        
\cline{2-6}                       
                        & GNN                   
                        & 85.00~[68.42,100.00]      
                        & 53.85~[34.78,72.74]      
                        & 58.62~[42.30,77.15]    
                        & 69.39~[54.44,83.33]
                        \\ 
                       
\cline{2-6}
                        & Transformer          
                        & 95.00~[83.33,100.00]      
                        & 38.46~[20.83,56.00] 
                        & 54.29~[37.50,70.97]
                        & 69.09~[53.33,81.48]
                        \\ 
\cline{2-6}
                        & Mamba                 
                        & 85.00~[68.18,100.00]
                        & 53.85~[34.78,73.69]
                        & 58.62~[40.90,76.67]      
                        & 69.39~[53.33,81.63] 
                        \\

\hline
\multirow{3}{*}{Text}   & Zero-shot LLM                                 
                        & 45.00~[23.53,68.19]  
                        & 96.15~[87.50,100.00]
                        & 90.00~[66.67,100.00]
                        & 60.00~[34.78,78.95]
                        \\ 
\cline{2-6}
                        & Few-shot LLM          
                        & 90.00~[73.32,100.00]    
                        & 61.54~[42.84,79.31]    
                        & 64.29~[45.71,80.65]      
                        & 75.00~[59.08,86.79] 
                        \\ 
                       
\cline{2-6}
                        & Few-shot LLM+Manual 
                        & 86.67~[66.67,100.00]     
                        & 72.73~[53.82,90.48]      
                        & 68.42~[46.67,88.00]      
                        & 76.47~[57.14,90.33]    
                        \\
\hline
\end{tabular}
}
\end{table}

\section{Discussion} \label{discussion}

In this research, the effectiveness of our proposed method for summarizing transcripts and predicting suicide risk using large language models is demonstrated, which outperforms traditional deep learning models, as clearly shown in Section~\ref{sec:results}. Specifically, our method achieved an F1 score of 76.47\%,a sensitivity of 86.68\%, a specificity of 72.73\%, and a precision of 68.42\%. Furthermore, the combination of LLMs with manual scales has shown significant performance in practical applications.These findings not only validate the feasibility of applying large models in the field of psychological suicide prediction but also underscore the importance of incorporating expert human intervention. Not only the predictive performance of LLM is remarkable, but also the integration with psychological operator scores from scales makes the prediction process more comprehensive and reliable, significantly enhancing the credibility of the method. 
% This comprehensive approach provides a solid foundation for future research and applications, potentially advancing the development and practical implementation of psychological suicide prediction methods.

From the examples shown in Figure~\ref{fig:example}, we observed that callers exhibited psychological issues arising from different situations, such as case~(A) with no specific stressor but have strong suicidal ideation, case~(B) related to adolescent challenges, case~(C) involving emotional problems, and case~(D) linked to work-life stress. Across these four examples, the general trend of assessments by both the manual scale and the AI model was similar, yet some nuanced differences in judgment revealed the strengths and limitations of each method.
In case~(A), although the caller did not display obvious symptoms of depression, they clearly articulated a method of suicide and showed significant emotional volatility. The manual scale rated this individual at 12/16, indicating a high suicide risk. Both the AI model and the manual scale rated the caller as being at high risk for suicide, and this was confirmed through follow-up when the caller did indeed attempt suicide. This consistency in judgment highlights the accuracy of both the AI and manual methods in identifying high-risk individuals.
In case~(B), the caller was a 13-year-old girl. Due to her adolescent status and the unique challenges she faced, she did not explicitly state a suicidal intent or plan. Consequently, both the operator and the AI model assessed her as being at medium to low risk, or as not being at risk of suicide. However, follow-up revealed that she eventually attempted suicide. This case~(D)raws attention to the need for specialized models and assessment criteria tailored specifically to adolescent issues, which we plan to develop in the next phase of our research.
In case~(C), the caller expressed intense stress from multiple sources and a strong desire to commit suicide. Both the operator and the AI model assessed the caller as being at high risk of suicide. Despite the fact that the caller did not attempt suicide during the follow-up period, we believe that such individuals still warrant close attention. Thus, the AI model's prediction is considered meaningful, as it rightly identified a situation that required significant concern.
In case~(D), the caller indicated multiple stressors related to work, finances, and family, and explicitly mentioned suicidal thoughts. The operator assessed this individual as being at medium to low risk, whereas the AI model, capturing key information, judged the caller as being at high risk of suicide. Follow-up confirmed that the AI model's assessment was accurate. This example demonstrates the value of the AI model, as it is capable of uncovering underlying information and effectively alerting operators to high-risk individuals.

In summary, these cases illustrate the potential of AI models in enhancing the accuracy of suicide risk assessments. While the manual scale and AI often align, the AI model's ability to detect subtle cues and assess high-risk situations accurately, especially in complex cases, underscores its importance in supporting mental health professionals. These findings support the continued development and refinement of AI-based tools to improve the identification and intervention for individuals at risk of suicide, with particular attention to special populations like adolescents.

Our current study still have several biases and limitations. Firstly, although our training set contains 1,238 long speeches, the test set includes only 46 long speeches, resulting in a relatively small sample size. In the future, we plan to expand the test set to achieve more reliable results. Secondly, the current model's efficiency in predicting callers' suicidal tendencies is inadequate, making it unable to provide real-time assistance to operators. We will continue to explore ways to better support operators in managing these situations. Thirdly, we have not yet analyzed how factors such as gender, age, and threatening events may impact the model's predictions, which could lead to biased results. In the future, we will further investigate the influence of these factors on the prediction of suicidal behavior.

In summary, the method we developed aims to leverage the advanced capabilities of artificial intelligence, particularly LLM, to assist psychological operators in accurately predicting suicide risk. We innovatively propose the integration of LLM into the field of suicide awareness detection for suicide assistance hotlines. This approach involves a deep analysis and comprehensive processing of psychological support hotline call records using LLMs. We conducted comparative experiments to highlight the performance differences between traditional natural language processing prediction models and LLMs, thereby underscoring the significant potential of the latter in suicide prevention tasks. Furthermore, we emphasize the role of human-computer interaction, combining LLMs with manually evaluated psychological suicide risk scales to achieve a more comprehensive and accurate prediction of suicide risk.

\section{Acknowledgments}\label{sec:acknowledgments}
This study was supported by the National Natural Science Foundation of China [82071546], Beijing Municipal High Rank Public Health Researcher Training Program [2022-2-027], the Beijing Hospitals Authority Clinical Medicine Development of Special Funding Support [ZYLX202130], and the Beijing Hospitals Authority’s Ascent Plan [DFL20221701]. Guanghui Fu is supported by the Chinese Government Scholarship provided by China Scholarship Council (CSC). The funding institutions had no role in the design, data collection, analysis, interpretation, writing of the report, or the decision to submit for publication.

\section{Ethical statement}
We prioritize participant privacy and are dedicated to safeguarding the security of their personal information, ensuring anonymity and confidentiality throughout the data processing phase. All analyses conducted in this study aim to advance scientific knowledge and promote social welfare, with no commercial or financial conflicts of interest. The research team assumes full responsibility for all content and conclusions presented in this study.

\newpage
\appendix
\renewcommand\thefigure{S\arabic{figure}} 
\setcounter{figure}{0} \renewcommand\thetable{S\arabic{table}} 
\setcounter{table}{0}   

\bibliography{refs} 
\bibliographystyle{spiebib} 
\end{document}